# Physics-Informed Super-Resolution Diffusion for 6D Phase Space Diagnostics

Alexander Scheinker[1,*]

[1]Applied Electrodynamics Group, Los Alamos National Laboratory, Los Alamos, New Mexico 87545, USA.
[*]ascheink@lanl.gov

Adaptive physics-informed super-resolution diffusion is developed for non-invasive virtual diagnostics of the 6D phase space density of charged particle beams. An adaptive variational autoencoder (VAE) embeds initial beam condition images and scalar measurements to a low-dimensional latent space from which a $32^6$ pixel 6D tensor representation of the beam's 6D phase space density is generated. Projecting from a 6D tensor generates physically consistent 2D projections. Physics-guided super-resolution diffusion transforms low-resolution images of the 6D density to high resolution 256×256 pixel images. Un-supervised adaptive latent space tuning enables tracking of time-varying beams without knowledge of time-varying initial conditions. The method is demonstrated with experimental data and multi-particle simulations at the HiRES UED. The general approach is applicable to a wide range of complex dynamic systems evolving in high-dimensional phase space. The method is shown to be robust to distribution shift without re-training.

## I. INTRODUCTION

Particle accelerators are incredibly flexible scientific tools for a wide range of research including material science, high energy physics, chemistry, and biology at extreme time scales. Free electron lasers (FEL) such as the Swiss FEL [1] and the FERMI FEL [2] use relativistic electrons to create bright flashes of highly coherent light for imaging non-crystalline objects such as single proteins and dynamic events such as shocks [3]. In ultrafast electron diffraction (UED) fs electron beams are used directly for imaging [4].

The European X-ray FEL at DESY accelerates nC electron bunches up to 17.5 GeV to generate femtosecond pulses of photons [5] and the Linac Coherent Light Source (LCLS) at SLAC can generate few fs long pulses with 12 keV photons utilizing 17 GeV electron bunches [6]. Beam-driven plasma wakefield acceleration (PWFA) typically relies on large accelerators with high energy beams such as the facility for advanced accelerator experimental tests (FACET) and FACET-II where 20 GeV fs-long bunches are generated with thousands of kA peak currents [7, 8], and at CERN's Advanced Proton Driven Plasma Wake-field Acceleration Experiment (AWAKE) facility with 400 GeV protons from CERN's Super Proton Synchrotron (SPS) accelerator to accelerate 18.8 MeV electron bunches up to 2 GeV [9]. Achieving such high beam energies with conventional radio frequency (RF) accelerator technology requires kilometer long accelerators. PWFA has the potential to enable smaller accelerators by creating GV/m accelerating gradients. Laser-driven PWFA (LWFA) accelerates electrons to high energy in an incredibly short distance, has been demonstrated for reliable production of fs X-ray beams with tunable polarization [10], and will potentially enable compact accelerators such as compact FELs [11].

All accelerator beams evolve in a 6D phase space $(x,y,z,p_x,p_y,p_z)$ subject to complex collective effects including space charge forces and coherent synchrotron radiation. In PWFA the complexities are even greater due to the complicated intense beam-plasma interactions [12, 13], for which custom shaping and control of a beam's 6D phase space is incredibly important. At light sources the beam's 6D phase space defines the properties of the generated light and must be adjusted quickly between various experiments. For beam-driven PWFA the beam's 6D phase space density defines the acceleration and energy spread achieved. For LWFA the generated beam's 6D phase space must be controlled for multi-staging and for achieving LWFA-based FELs. Precise control of a beam's 6D phase space requires an ability to measure it.

This paper develops a general adaptive and physics-informed method for virtual beam diagnostics for tracking the 6D phase space of time-varying beams by incorporating adaptive feedback within a variational autoencoder (VAE) and a super-resolution diffusion model. The VAE generates 6D tensor representations of a beam's phase space density from which all 15 unique 2D views are created as projections of a single tensor. The accuracy of those images is increased by a physics-informed conditional guided super-resolution diffusion process. The adaptive process is demonstrated with experimental data from the HiRES UED at Lawrence Berkeley National Laboratory shown in Figure 1. Figure 2 shows an overview of the generative approach. This general method is applicable to a wide range of complex dynamic systems evolving in high-dimensional phase space.

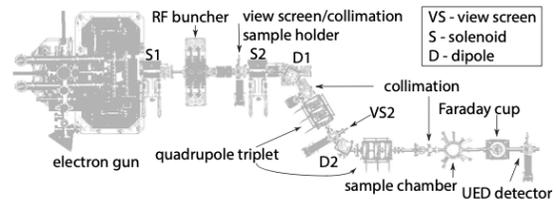

FIG. 1. High level overview of the HiRES UED.

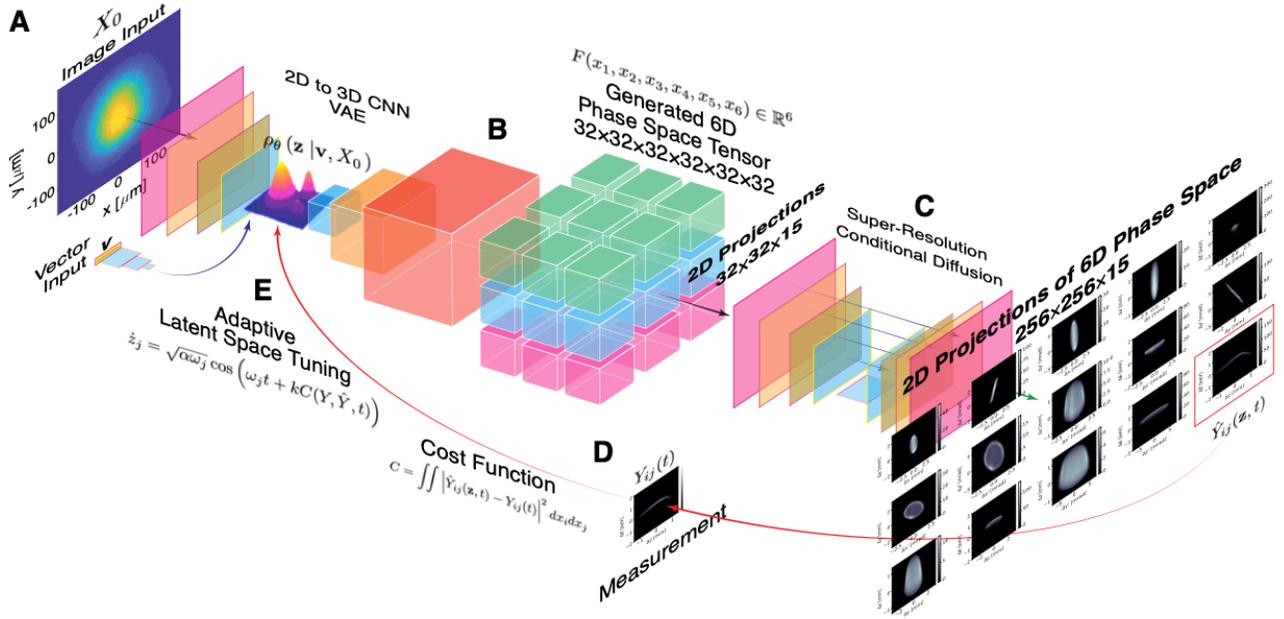

FIG. 2. Overview of the super-resolution 6D diffusion approach. **A**: A variational autoencoder (VAE) embeds a measurement of the beam's initial $(x, y)$ distribution at the accelerator entrance together with a vector **v** which contains bunch charge and solenoid setting into a latent vector $\mathbf{z} \in \mathbb{R}^3$. **B**: The decoder half of the VAE generates a $32^6$ pixel 6D tensor object which is then projected down to all 15 unique 2D combinations to form all of the 15 unique 2D projections of the beam's 6D phase space. **C**: The resolution of each $32 \times 32$ pixel projection is increased to $256 \times 256$ pixels by using a super-resolution diffusion process. **D**: One of the generated projections is compared to its single-shot measurement and the error defines a cost function. **E**: The cost function is minimized by adaptively tuning the low-dimensional latent space embedding of the VAE.

## II. OVERVIEW OF MACHINE LEARNING FOR ACCELERATOR APPLICATIONS

Recently, the world's first 6D beam phase space measurement was achieved at the Spallation Neutron Source Beam Test Facility [14]. The 6D phase space measurement process is not a single-shot reconstruction, it currently requires approximately 18 hours in which millions of projections of distinct beams are measured under stable conditions in which beam to beam fluctuations are minimal.

Enabling real-time 6D phase space control requires much faster techniques to measure a beam's 6D density. Machine learning (ML) based methods have the potential to significantly speed up such measurements. ML tools are being developed to provide high accuracy phase space diagnostics of particle accelerator beams [15], for many-body interactions [16], for lattice quantum Monte Carlo simulations [17], for quantum dynamics [18], for 3D reconstructions of the electron density of crystals for coherent diffraction imaging [19], for quantum feedback [20], and for determining the structures of unknown networks with time delays [21].

There are many efforts underway to develop ML-based tools for laser plasma physics in general [22], and specifically for LWFA optimization and tuning [23, 24], development of tools that can act as virtual diagnostics providing information about a beam's 6D phase space including adaptive latent space tuning of generative autoencoders for tracking time-varying beams [25, 26], normalizing flows [27], and frameworks that combine generative ML models together with physics models which are optimized and tuned based on experimental beam measurements [28], maximum entropy methods [29]. Conditional guidance for generative models has also been explored in a recent novel approach of utilizing conditionally guided neural networks for errant beam [30]. A group at Fermilab is currently developing a general ML operations framework for accelerator control [31]. Researchers from Pacific Northwest National Laboratory (PNNL) and Fermilab are collaborating on developing methods for automatically identifying causes of beam times by using advanced AI/ML techniques such as LSTMs, GANs, and transformers [32]. A group from Los Alamos National Laboratory (LANL) has developed generative AI models for electrodynamics with hard physics constraints [33], and adaptive ML methods which combine deep neural networks with model-independent adaptive extremum seeking feedback [34], which have been demonstrated for automatic control of the (z,E) longitudinal phase space distribution of electron beams in SLAC's LCLS FEL [35]. SLAC researchers have developed Bayesian methods that have been demonstrated for real-time tuning of FELs [36].

Neural networks are also being used for uncertainty aware anomaly detection to predict errant beam pulses [37], as virtual diagnostics for 4D tomographic phase space reconstructions [38], for predicting the transverse emittance of space charge dominated beams [39], and for magnet control [40]. At the SwissFEL, Bayesian methods with safety constraints are being developed [41], and at the EuXFEL, CNNs have been used to generate high resolution longitudinal phase space diagnostics [15]. At the Central Laser Facility, Gaussian processes have been used for LWFA optimization [24].

## III. BEAM DYNAMICS

Charged particle dynamics take place in a 6D position momentum $(\mathbf{r}, \mathbf{p})$ phase space

$$\mathbf{r} = (x, y, z), \qquad \mathbf{p} = \gamma m \mathbf{v}, \quad \gamma = 1/\sqrt{1 - v^2/c^2}, \quad (1)$$

where $\gamma$ is the relativistic Lorentz factor, $v = dr/dt$, and $v = \|\mathbf{v}\|$. The evolution of the 6D phase space density $X(r,p,t)$ of a large collection of charged particles in a beam can be described by the relativistic Vlasov equation

$$\frac{\partial \mathbf{X}}{\partial t} + \mathbf{v} \cdot \nabla_\mathbf{r} \mathbf{X} + \frac{\partial \mathbf{p}}{\partial t} \cdot \nabla_\mathbf{p} \mathbf{X}, \quad \frac{d\mathbf{p}}{dt} = q(\mathbf{E} + \mathbf{v} \times \mathbf{B}), \quad (2)$$

where the electric and magnetic fields can each be separated into external and beam-based components

$$\mathbf{E}(t) = \mathbf{E}_{ext}(t) + \mathbf{E}_{beam}(t), \quad \mathbf{B}(t) = \mathbf{B}_{ext}(t) + \mathbf{B}_{beam}(t).$$

The external parts of the fields are generated by accelerator components such as magnets and resonant accelerating structures. The beam-based components are those due to the charges and electric fields of the beams them- selves and can be calculated from the beam's charge and current density which are defined as

$$\rho(\mathbf{r}, t) = \iiint \mathbf{X}(\mathbf{r}, \mathbf{p}, t) d\mathbf{p}, \quad \mathbf{J}(\mathbf{r}, t) = \mathbf{v}(t)\rho(\mathbf{r}, t). \quad (3)$$

The self-fields of the beam satisfy Maxwell's equations

$$\nabla_\mathbf{r} \cdot \mathbf{E}_{beam}(\mathbf{r}, t) = \frac{\rho(\mathbf{r}, t)}{\epsilon_0},$$

$$\nabla_\mathbf{r} \times \mathbf{B}_{beam}(\mathbf{r}, t) = \mu_0 \left( \mathbf{J}(\mathbf{r}, t) + \epsilon_0 \frac{\partial \mathbf{E}_{beam}(\mathbf{r}, t)}{\partial t} \right) \quad (4)$$

To simulate beam dynamics (2)-(4) for realistic beams represented by hundreds of millions of macro particles is computationally expensive. Recent efforts have focused on creating generative AI-based methods with hard physics constraints to speed up such calculations, but they are not yet fully developed [33]. Even when fast GPU-based simulations exist the initial conditions of the beams and of the accelerator components are uncertain and time-varying. While at higher energies the beam can be passed through screens for single shot (x, y) measurements, measuring initial low energy beam conditions can be more difficult as screens at low energy have a large in- fluence on the beam's characteristics and other processes are also slow and invasive such as wire scans or quad scans that interrupt the beam. While some of the accelerator components can be measured in real-time even these measurements suffer from arbitrary drifts and noise. In this work, we develop generative AI-based methods for predicting the beam's 6D phase space at downstream accelerator locations even as the beam's initial conditions and accelerator parameters vary with time and are not available for measurement.

## IV. VARIATIONAL AUTOENCODER FOR 6D TENSOR GENERATION

The HiRES UED accelerates pC-class, sub-picosecond long electron bunches at MHz rates, providing some of the most dense 6D phase space among accelerators at unique repetition rates [42, 43].

The first part of our approach is the development of a variational autoencoder (VAE) which can embed a combination of HiRES beam initial conditions and accelerator parameter measurements into a low-dimensional latent space vector $z \in R^3$ [44]. Creating a low-dimensional representation of the beam initial conditions and accelerator parameters enables fast adaptive tuning of this representation once the generative model has been trained and these quantities start to vary with time but are no longer available for measurement.

Here we note that while we described the 6D phase space coordinates above as $(x,y,z,p_x,p_y,p_z)$, in typical accelerator applications the beam is traveling down the z-axis with very high energy so that the longitudinal momentum component is orders of magnitude greater than the transverse momentum components: $p_z \gg p_x, p_y$. Our typical measurement of interest is actually the transverse angles at which the particle are moving relative to the z- axis, which we denote by $x' = p_x/p_z$ and $y' = p_y/p_z$. Furthermore, we typically assume that almost all of the kinetic energy is coming from $p_z$ and simply refer to the total beam energy as E. For the remainder of the paper we then focus on the following 6D phase space coordi- nates $(x, y, z, x', y', E)$.

We consider the time-varying initial 6D phase space density of the beam at the accelerator entrance (z = 0), which we denote by $X^0(t)$ for which we have collected experimental measurements of the (x, y) projection

$$\mathbf{X}^0_{x,y}(t) = \Pi^{x,y} \mathbf{X}^0(t) = \iiiint \mathbf{X}^0(t) dx' dy' dz dE + n(t), \quad (5)$$

where n(t) is measurement noise. We also consider a vector of measurable accelerator and beam parameters p(t) = [q(t), s(t)], where q(t) represents the time-varying charge of the beams generated at the HiRES photocathode and s(t) represents a time-varying solenoid current of S1 in Figure 1.

A high-level overview of the VAE architecture developed for this work is shown in Figure 3. The experimentally measured beam image $X^0_{x,y}$ of size 52 × 52 pixels passes through pairs of 2D convolutional layers which utilize leaky ReLU activation functions as f(x)=0.1x for x<0 and x for x>=0. After 2 convolutional layers a maxpool layer over a 2 × 2 grid is applied which halves the image size with zero padding at the edges. After 3 such image steps the original 52 × 52 pixel image is reduced to 6 × 6 pixels before being flattened and passed through a dense layer. In parallel the vector of parameters p is passed through several dense layers and then concatenated with the output of the dense layer from the image branch of the encoder network. The resulting vector is then split into two dense networks each of which consists of two layers of 128 weights each with the same leaky ReLU activation functions before passing through linear dense layers of size 3. The output of the VAE encoder is 2 vectors, one that represents a mean value $M_\theta(X^0_{x,y})$ and another that represents a diagonal covariance matrix $\Sigma_\theta(X^0_{x,y}) \in \mathbb{R}^{3\times 3}$, where $\theta$ represents the weights of the encoder neural network. A random vector $\epsilon \in \mathbb{R}^3$ is then sampled from a mean-zero unit variance normal distribution N (0, I) and the latent vector representation is constructed as

$$\mathbf{z} = F_{en}(\mathbf{X}, \sigma) = M_\theta + \Sigma_\theta \epsilon \in \mathbb{R}^3, \quad (6)$$

which is then passed through the decoder part of the VAE. At this point this can be thought of as a regular auto-encoder whose latent embedding is perturbed with random noise before being passed to the decoder, but in this approach the encoder has the flexibility to learn the standard deviation of the noise and the latent representation is pushed towards a mean zero unit variance Gaussian distribution with a diagonal covariance matrix by minimization of the Kullback-Leibler (KL) divergence between the learned latent space probability density and the Gaussian distribution defined as

$$\mathcal{KL}(\theta, \mathbf{X}) = \frac{1}{2}\left(Tr[\Sigma_\theta] + M_\theta^T M_\theta - \log[\det(\Sigma_\theta)]\right). \quad (7)$$

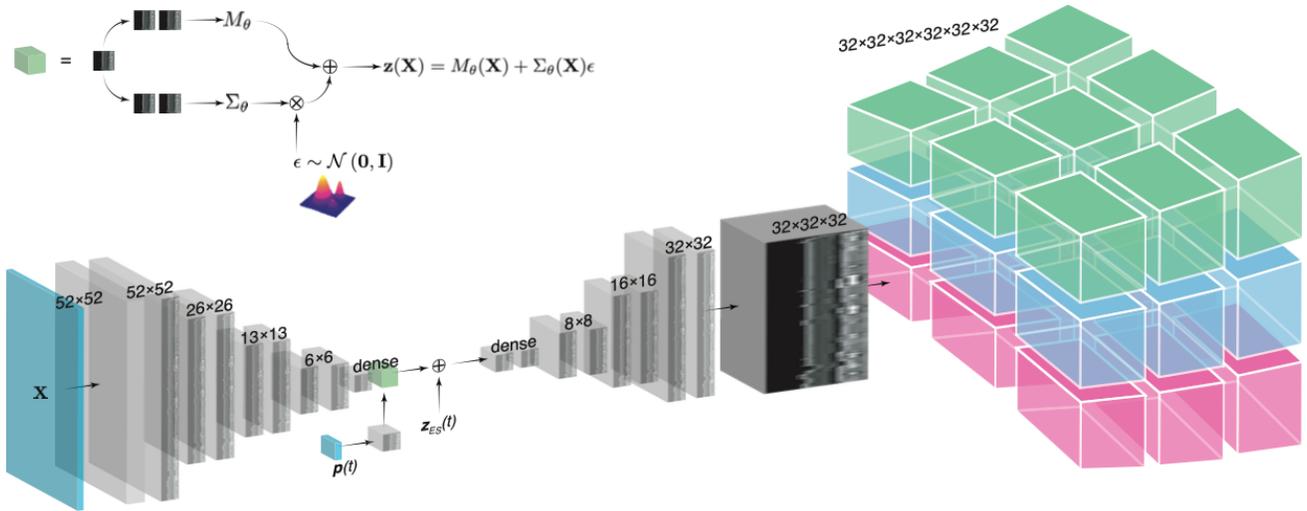

FIG. 3. An overview of the VAE architecture is shown. The $52 \times 52$ pixel input beam measurement images $\mathbf{X}^0_{x,y}(t)$ are compressed into latent embeddings in $\mathbb{R}^3$ and concatenates them with dense layers of the same size which act on the input parameters $\mathbf{p}(t)$. The latent embedding is added to an external perturbation signal $\mathbf{z}_{ES}(t)$ that enables adaptive tuning before passing through the decoder half of the network that eventually builds up a $32^6$ pixel representation.

An additional input is brought in at this stage, a vector $\mathbf{z}_{ES} \in \mathbb{R}^3$, which is kept at a constant value of 0 during training, but will be used later to adaptively tune the latent vector. The latent embedding vector is initially passed through a dense layer in the decoder and reshaped into a $8 \times 8$ pixel image that passes through a series of 2D convolutional layers, leaky ReLU activations, and transposed convolutional layers until an $32 \times 32$ pixel image with $32^2$ channels is created, which is then reshaped into a $32^3$ 3D tensor with 32 channels of size $32 \times 32 \times 32 \times 32$. After passing through two 3D convolutional layers, the number channels is increased to $32^3$ and finally a 6D tensor $\hat{X}^L_{lr}$ of shape $32\times32\times32\times32\times32\times32$ is generated which passes through a ReLU activation to enforce the physical constraint that the generated 6D density is positive. The overall VAE input-output flow is summarized as

$$\begin{aligned}[\mathbf{X}^0_{x,y}(t), \mathbf{p}(t)] &\longrightarrow \mathbf{z}(t) = F_{en}\left[\mathbf{X}^0_{x,y}(t), \mathbf{p}(t), \theta\right], \\ \mathbf{z}(t) &\longrightarrow \hat{\mathbf{X}}^L_{lr}(t) = F_{de}[\mathbf{z}(t), \varphi],\end{aligned} \quad (8)$$

where $\theta$ and $\varphi$ represent the encoder and decoder weights, respectively and the subscript lr emphasizes the fact that a low resolution $32^6$ object has been created at this point. The reason for this low resolution is because we are working with such a high dimensional object. When training with $32^6$-sized floating point 32 tensors the network is limited to a batch size of 1 even with a powerful NVIDIA A100 GPU with 80 GB of VRAM.

This 6D tensor $\hat{X}^L_{lr}$ is a generated approximation of the beam's 6D phase space density $X^L$ at a downstream location z = L. All 15 unique 2D projections (x,y), (x, x'), ..., (z, E) of $\hat{X}^L_{lr}$ are then generated according to projection operators

$$\hat{\mathbf{X}}^L_{lr,i_1 i_2} = \Pi^{i_1 i_2} \hat{\mathbf{X}}^L_{lr} = \iiiint \hat{\mathbf{X}}^L_{lr} di_3 di_4 di_5 di_6, \quad (9)$$

Where $i_j \in \{x,y,z,x',y',E\}$.

By projecting from a 6D tensor important physics constraints are hard-coded into VAE, such as charge conservation and physically consistent projections, which are otherwise not conserved with generative methods. Because the 6D representation is computationally expensive (a $32^6$ representation consists of approximately 1 billion pixels), the 15 projections are created as low- resolution 32 × 32 pixel images which require transformation into high-resolution 256 × 256 pixel images by a super-resolution diffusion process. Because of the low- dimensional latent embedding $z \in R^3$, we are able to quickly and adaptively modify the entire 6D distribution $\hat{X}^L_{lr}$ by adjusting only 3 numbers.

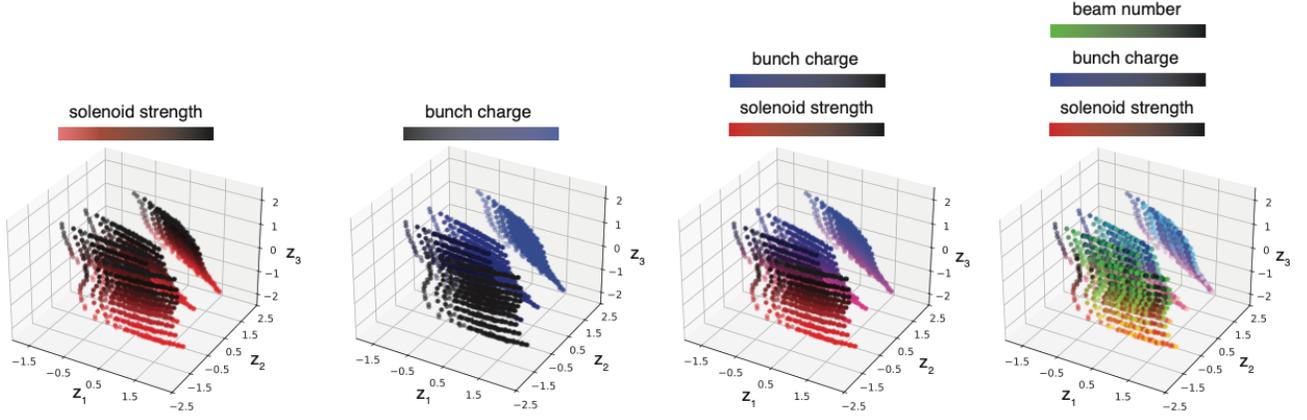

FIG. 4. Overview of the latent space embedding arrangement of beams with different initial conditions (labeled as beam number), solenoid settings, and bunch charge. In the first image from left to right only the solenoid strengths are colored as different shades of red. In the second from left image the latent positions are colored in shades of blue based on bunch charge. In the third image from left the red and blue channels are defined by solenoid strength and bunch charge, respectively. In the right most image each latent position is RGB-colored according to solenoid strength, bunch charge, and an additional green channel for input beam number.

The VAE is trained by using a single cost function which is a combination of the KL divergence together with the negative log likelihood of generating the true training data, resulting in the training cost

$$C = w \left(Tr[\Sigma_\theta(\mathbf{X})] + M_\theta^T(\mathbf{X})M_\theta(\mathbf{X}) - \log[\det(\Sigma_\theta(\mathbf{X}))]\right) \\ - \log\left[Pr(\mathbf{X}|\mathbf{z}(\mathbf{X}), \varphi)\right], \quad (10)$$

which is averaged over batches of data during training and where w > 0 is a small weight (we used 1e-3) which allows the latent embedding to arrange itself in a physically meaningful way rather than converging directly to a mean 0 normal distribution. This physically interpretable latent space arrangement is evident in Figure 4 where we see that $z_3$ corresponds with solenoid strength, an arc along $z_1$ corresponds to bunch charge, and various random initial conditions for a given solenoid strength or bunch charge are scattered randomly near their associated charge and magnet current latent space positions.

## V. SUPER-RESOLUTION DIFFUSION

The next step of this process is the development of a super-resolution diffusion model that can convert the coarse 32×32 pixel images up to high-resolution 256×256 pixel images. In the broad ML community, diffusion- based models are the state-of-the-art method for generative highly accurate representations of high dimensional complex data, such as high resolution images. Recently, the first application of diffusion was demonstrated as a virtual diagnostic to generate megapixel resolution virtual views of the (z,E) longitudinal phase space of the electron beam in the European X-ray FEL [45]. That method was then generalized to an adaptively tuned conditional diffusion approach for generating high resolution representations of all of the unique 2D projections of the HiRES UED beam's 6D phase space distribution [46].

Generative diffusion models are based on a gradual de- noising approach inspired by stochastic differential equations and statistical thermodynamics [47]. Diffusion is the state-of-the-art for the generation of high resolution images [48–51], especially when the images have a wide variety. Generative diffusion-based models are being developed for a wide range of scientific applications [52], such as conditional generation of hypothetical new families of superconductors [53], for brain imaging [54], for various bioengineering applications [55], for protein structure generation [56].

The first step of the diffusion approach is to create a noise schedule with which the training images are slowly transformed into signals indistinguishable from pure noise. Given an image, x, the diffusion process proceeds as defined by a discrete time version of a stochastic differential equation over time steps t = 0, 1, . . . , T according to

$$\begin{aligned}
\mathbf{z}_1 &= \sqrt{1-\beta_1}\mathbf{x} + \sqrt{\beta_1}\epsilon_1, \\
\mathbf{z}_2 &= \sqrt{1-\beta_2}\mathbf{z}_1 + \sqrt{\beta_2}\epsilon_2, \\
&\vdots \\
\mathbf{z}_t &= \sqrt{1-\beta_t}\mathbf{z}_{t-1} + \sqrt{\beta_t}\epsilon_t, \\
&\vdots \\
\mathbf{z}_T &= \sqrt{1-\beta_T}\mathbf{z}_{T-1} + \sqrt{\beta_T}\epsilon_T, \quad \epsilon_t \sim \mathcal{N}(\mathbf{0},\mathbf{I}),
\end{aligned} \quad (11)$$

where a nonlinear sinusoidal noise schedule is used as was first proposed in [50]. Here at first $\beta_1 = 1e-4$ and ends with $\beta_T = 1e-2$ with the $\beta_i$ schedule defined as

$$\beta_t = \beta_1 + (\beta_T - \beta_1) \times \sin\left(\frac{\pi i}{2T}\right), \quad t = 1, 2, \ldots, T. \quad (12)$$

Note that in our discussion of the noising and denoising diffusion process we use the standard convention of labeling iteration steps as $t=1,\ldots,T$ which is the diffusion time and un-related to the physical time scale on which our adaptive algorithm works with a time-varying beam $\mathbf{X}(t)$. This abuse of notation with a dual use of t is limited to just this section which describes the theory behind the generative diffusion profess. Everywhere else in the manuscript t refers to a physical time scale which describes the evolution of the time-varying accelerator and its beam.

The diffusion forward noising process as described by Equation (11) results in conditional distributions

$$\begin{aligned}
q(\mathbf{z}_1|\mathbf{x}) &= \mathcal{N}_{\mathbf{z}_1}\left(\sqrt{1-\beta_1}\mathbf{x}, \beta_1\mathbf{I}\right) \\
&\vdots \\
q(\mathbf{z}_t|\mathbf{z}_{t-1}) &= \mathcal{N}_{\mathbf{z}_{t-1}}\left(\sqrt{1-\beta_t}\mathbf{z}_{t-1}, \beta_t\mathbf{I}\right).
\end{aligned} \quad (13)$$

This Markov process has a joint distribution

$$q(\mathbf{z}_{1,2,\ldots,t}|\mathbf{x}) = q(\mathbf{z}_1|\mathbf{x}) \prod_{s=2}^{t} q(\mathbf{z}_s|\mathbf{z}_{s-1}), \quad (14)$$

which allows for a simple calculation of
.
$$q(\mathbf{z}_t|\mathbf{x}) = \mathcal{N}_{\mathbf{z}_t}\left(\sqrt{\alpha_t}\mathbf{x}, (1-\alpha_t)\mathbf{I}\right), \quad \alpha_t = \prod_{s=1}^{t}(1-\beta_s), \quad (15)$$

which allows us to easily calculate $z_t$ at any diffusion step $t$ directly from x for model training according to

$$\mathbf{z}_t = \sqrt{\alpha_t}\mathbf{x} + \sqrt{1-\alpha_t}\epsilon, \quad \epsilon \sim \mathcal{N}(\mathbf{0},\mathbf{I}). \quad (16)$$

Note that we choose $0 < \beta_t < 1$ such that

$$\lim_{t\to\infty}(\alpha_t, 1-\alpha_t) = (0,1).$$

In practice, with the choice of $\beta_t$ described above and T = 1000, $\alpha_T \approx 0$ and so $z_T$ converges very closely to pure noise N(0,I), therefore in what follows we assume that

$$P(\mathbf{z}_T) = \mathcal{N}_{\mathbf{z}_T}(\mathbf{0},\mathbf{I}). \quad (17)$$

At this step we make a choice that is similar to what is done when building a VAE, in order to learn the reverse diffusion process, to create images, we assume that we can model the process as

$$P(\mathbf{z}_{t-1}|\mathbf{z}_t, \varphi_t) = \mathcal{N}_{\mathbf{z}_{t-1}}\left(\mathcal{D}_t[\mathbf{z}_t,\varphi_t], \sigma_t^2\mathbf{I}\right), \quad (18)$$

where $\varphi_t$ are the time-dependent components of the diffusion neural network D. The model weights can then be found by the usual method of minimizing the negative log-likelihood of the training data

$$-\sum_i \log\left[P(\mathbf{x}_i|\varphi_t)\right]. \quad (19)$$

Unfortunately, calculating the integral

$$P(\mathbf{x}_i|\varphi_t) = \int P(\mathbf{x},\mathbf{z}_{1,\ldots,T}|\varphi_{1,\ldots,T})d\mathbf{z}_{1,\ldots,T} \quad (20)$$

is intractable, so the problem is transformed by applying Jensen's inequality and working with the evidence lower bound (ELBO) loss which satisfies the inequality

$$\begin{aligned}
&\log\left[P(\mathbf{x}_i|\varphi_t)\right] \\
&= \log\left[\int P(\mathbf{x},\mathbf{z}_{1,\ldots,T}|\varphi_{1,\ldots,T})d\mathbf{z}_{1,\ldots,T}\right] \\
&\geq \underbrace{\int q(\mathbf{z}_{1,\ldots,T})\log\left[\frac{P(\mathbf{x},\mathbf{z}_{1,\ldots,T}|\varphi_{1,\ldots,T})}{q(\mathbf{z}_{1,\ldots,T}|\mathbf{x})}\right]d\mathbf{z}_{1,\ldots,T}}_{\text{ELBO}} \quad (21)
\end{aligned}$$

Therefore the negative log-likelihood is strictly less than the negative ELBO loss and so minimizing the negative ELBO loss also pushes down the negative log- likelihood. The ELBO loss itself can be simplified by application of Equations (11), (18), and Bayes' rule, and the recognizing terms that are in the form of the KL divergence discussed for VAE training, resulting in

$$\begin{aligned}
\text{ELBO} &= \log\left[\mathcal{N}_{\mathbf{x}}\left(\mathcal{D}_1[\mathbf{z}_1,\varphi_1], \sigma_1^2\mathbf{I}\right)\right] \\
&- \sum_{t=2}^{T}\mathbb{E}_{q(\mathbf{z}_t|\mathbf{x})}\left[D_{\text{KL}}\left(q(\mathbf{z}_{t-1}|\mathbf{z}_t,\mathbf{x})|\mathcal{N}_{\mathbf{z}_{t-1}}\left(\mathcal{D}_t[\mathbf{z}_t,\varphi_t],\sigma_t^2\mathbf{I}\right)\right)\right]
\end{aligned}$$

where we need to write out the form of the conditional distribution $q(z_{t-1}|z_t,x)$ for training, which can be done by using Bayes' rule, applying the forward diffusion Equation (11), and performing a Gaussian change of variables as well as the Gaussian identity that relates the products of two normal distributions to a new normal distribution in terms of the means and standard deviations of the original two distributions. We end up with $q(z_{t-1}|z_t, x)$ proportional to

$$\mathcal{N}_{\mathbf{z}_{t-1}}\left[\frac{1-\alpha_{t-1}}{1-\alpha_t}\sqrt{1-\beta_t}\mathbf{z}_t + \frac{\sqrt{\alpha_{t-1}}}{1-\alpha_t}\beta_t\mathbf{x}, \frac{1-\alpha_{t-1}}{1-\alpha_t}\beta_t\mathbf{I}\right]. \quad (22)$$

The KL divergence between the two normal distributions $q(z_{t-1}|z_t,x)$ and $N\,D_t[z_t,\varphi_t],\sigma_t^2 I$ in the ELBO

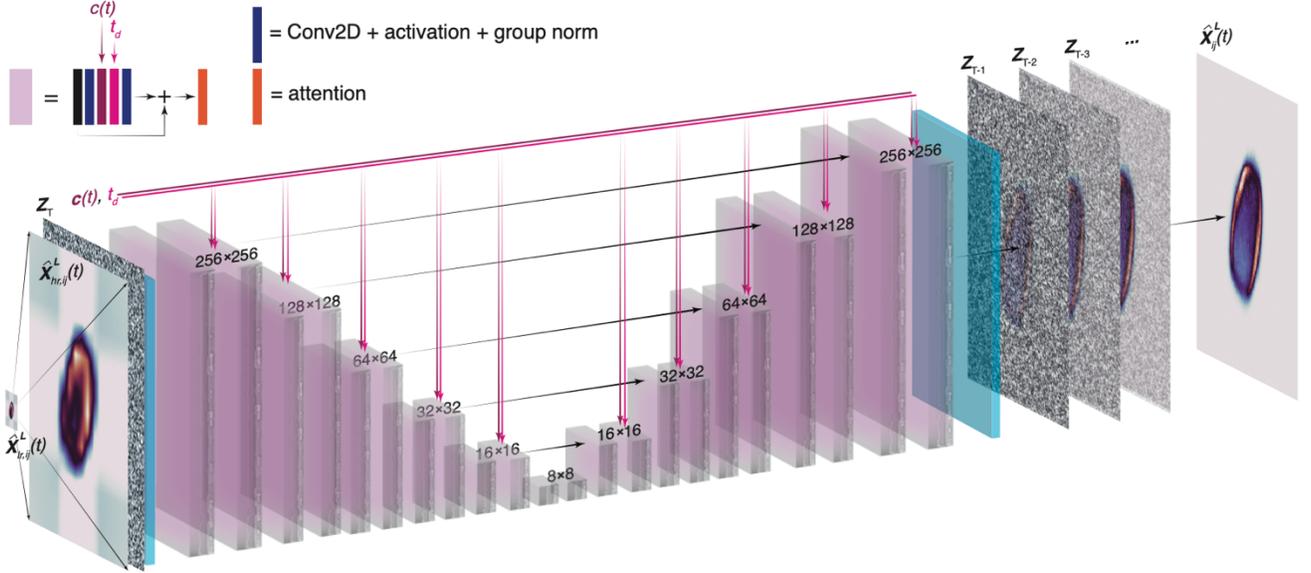

FIG. 5. An overview of the UNet architecture used for the super resolution diffusion process is shown.

term can then be expressed analytically (neglecting additive constants that do not influence optimization) as

$$\frac{1}{2\sigma_t^2} \left\| \frac{1-\alpha_{t-1}}{1-\alpha_t}\sqrt{1-\beta_t}\mathbf{z}_t + \frac{\sqrt{\alpha_{t-1}}}{1-\alpha_t}\beta_t\mathbf{x} - \mathcal{D}_t[\mathbf{z}_t,\varphi_t] \right\|^2 \quad (23)$$

The total negative ELBO loss that should be minimized is then given by averaging the sum of

$$-\log\left[\mathcal{N}_\mathbf{x}\left(\mathcal{D}_1[\mathbf{z}_1,\varphi_1],\sigma_1^2\mathbf{I}\right)\right]$$

and

$$\sum_{t=2}^{T} \frac{1}{2\sigma_t^2} \left\| \frac{1-\alpha_{t-1}}{1-\alpha_t}\sqrt{1-\beta_t}\mathbf{z}_t + \frac{\sqrt{\alpha_{t-1}}}{1-\alpha_t}\beta_t\mathbf{x} - \mathcal{D}_t[\mathbf{z}_t,\varphi_t] \right\|^2$$

over batches of training data at various diffusion times. In practice, it was found that instead of taking $(t,z_t)$ as inputs and trying to predict at each diffusion time t the slightly less noisy image $z_{t-1}$ directly via $D_t[z_t,\varphi_t]$, the model works better if it instead tries, at each diffusion time t to predict the noise term $\varepsilon_t$ that was mixed with the original x in order to create $z_t$. This approach makes the problem easier for the diffusion network as it is now solving the problem of determining a small residual rather than generating an entire image from scratch. The iterative generative diffusion process then works by taking in $z_T$, predicting the noise $\varepsilon_T$, removing that noise from $z_T$ to create an estimate of $z_{T-1}$, and continuing in this way iteratively until finally reconstructing $z_1$ and then x. This allows the model to take advantage of our knowledge of the noising process, specifically this approach takes advantage of Equation 16 which allows us
to approximate x relative to an approximation of $z_t$ as

$$\mathbf{x} = \frac{1}{\alpha_t}\mathbf{z}_t - \frac{\sqrt{1-\alpha_t}}{\sqrt{\alpha_t}}\epsilon, \quad (24)$$

which is substituted into the sum in the negative ELBO loss equation allows us to rewrite it as

$$\sum_{t=2}^{T} \frac{1}{2\sigma_t^2} \left\| \frac{1}{\sqrt{1-\beta_t}}\mathbf{z}_t - \frac{\beta_t}{\sqrt{1-\alpha_t}\sqrt{1-\beta_t}}\epsilon_t - \mathcal{D}_t[\mathbf{z}_t,\varphi_t] \right\|^2.$$

During inference, when generating images, we start with $z_T \sim N(0,I)$ and start to work backwards, therefore at each generative diffusion time step t, when we are working to generate $z_{t-1}$, we have already created our estimate of $z_t$ in the previous step. Therefore, in the the equation above we assume that we know $z_t$ and thereby simplify the problem being solved by the diffusion model by rewriting its output as

$$\mathcal{D}_t[\mathbf{z}_t,\varphi_t] = \frac{1}{\sqrt{1-\beta_t}}\mathbf{z}_t - \frac{\beta_t}{\sqrt{1-\alpha_t}\sqrt{1-\beta_t}}\mathcal{D}_{\epsilon,t}[\mathbf{z}_t,\varphi_t], \quad (25)$$

and the model only has to predict the error term $\varepsilon_t$. Plugging in this form of $D_t$ the entire negative ELBO loss equation simplifies to minimizing the mean squared error

$$\sum_{t=1}^{T} \frac{\beta_t^2}{(1-\alpha_t)(1-\beta_t)2\sigma_t^2} \left\|\mathcal{D}_{\epsilon,t}[\mathbf{z}_t,\varphi_t] - \epsilon_t\right\|^2. \quad (26)$$

In our application, all of the above analysis stays exactly the same, but our diffusion model has two additional inputs. The first additional input is a conditional vector **c** which informs the diffusion model which projection should be generated at which accelerator settings. The second conditional input is a high resolution ($256 \times 256$ pixels) image $\hat{\mathbf{X}}_{hr,ij}^L$ version of the low resolution ($32 \times 32$ pixels) image $\hat{\mathbf{X}}_{lr,ij}^L$ that was generated by the VAE for which we would like to perform super-resolution, where the high resolution version has been created by a simple bilinear interpolation, which is concatenated with the noisy image $\mathbf{z}_t$. Therefore, in our approach the diffusion model's arguments are

$$\mathcal{D}_{\epsilon,t}\left[\left(\mathbf{z}_t, \hat{\mathbf{X}}_{hr,ij}^L\right), \mathbf{c}, \varphi_t, t\right]. \quad (27)$$

The conditional vector consists of two parts, $\mathbf{c} = [\mathbf{d}, \mathbf{z}]$, where $\mathbf{z} \in \mathbb{R}^3$ is the latent embedding generated by the VAE which summarizes the physical state of the accelerator and the initial condition of the beam at the accelerator entrance. The vector **d** is a physics-informed vector that specifies which of the 15 projections of the beam's 6D phase space needs to be being generated. Naively, one could design **d** as just a normalized scalar value $d \in \mathbb{R}$ uniformly spread over 15 steps between the values 0 and 1, but this arbitrarily places the representation of certain projections closer together than others. For example, if the projections are ordered as $(x,y), (x,z), (x,E), (y,y'), \ldots$, then representing them with conditional numbers $0, 1/14, 2/14, 3/14, \ldots$ results in representation of $(x,y)$ and $(x,E)$ being $2/4$ apart while $(x,E)$ and $(y,y')$ are twice as close together at $1/14$ despite the fact that $(x,E)$ and $(x,y)$ share one axis while $(x,E)$ and $(y,y')$ are independent. Such arbitrary conditional inputs make it difficult for the network to take advantage of the physical relationships between the various projections.

A natural alternative to this, based on the standard approach taken in classification problems, is to simply use a one-hot encoding such that $\mathbf{d} \in \mathbb{R}^{15}$ has all 0 entries except for the position of the single projection out of the 15 that should be generated. Continuing with the above example, this would represent $(x,y)$ as $\mathbf{d}_{x,y} = [100\ldots]$, $(x,z)$ as $\mathbf{d}_{x,z} = [010\ldots]$, $(x,E)$ as $\mathbf{d}_{x,E} = [0010\ldots]$, and $(y,y')$ as $\mathbf{d}_{y,y'} = [00010\ldots]$. In this approach all vector are orthogonal and equidistant $\|\mathbf{d}_i - \mathbf{d}_j\|_2 = \sqrt{2}$. This is a better approach but it ignores some of the relationships between the projections.

Our study found that the best approach is to use a physics informed approach similar to one-hot encoding in which we use vectors $\mathbf{d} \in \mathbb{R}^6$ with 1 entries corresponding to the two phase space dimensions that are present in the projection. In the above example, this would represent $(x,y)$ as $\mathbf{d}_{x,y} = [1100\ldots]$, $(x,z)$ as $\mathbf{d}_{x,z} = [1010\ldots]$, $(x,E)$ as $\mathbf{d}_{x,E} = [1001\ldots]$, and $(y,y')$ as $\mathbf{d}_{y,y'} = [010001\ldots]$ so that $\mathbf{d}_{x,y} \cdot \mathbf{d}_{x,y} = 1$ and $\|\mathbf{d}_{x,y} - \mathbf{d}_{x,E}\|_2 = \sqrt{2}$, while $\mathbf{d}_{y,y'} \cdot \mathbf{d}_{x,E} = 0$ and $\|\mathbf{d}_{y,y'} - \mathbf{d}_{x,E}\|_2 = 2$. This incorporates the physical relationships between the various projections, making only the independent ones orthogonal and further away from each other than those which share an axis. This overall design is summarized in Table I.

TABLE I. Conditional vectors $\mathbf{d}_{ij}$ for $i, j \in \{x, y, z, x', y', E\}$.

| x | y | z | x' | y' | |
|---|---|---|---|---|---|
| [110000] | | | | | y |
| [101000] | [011000] | | | | z |
| [100100] | [010100] | [001100] | | | x' |
| [100010] | [010010] | [001010] | [000110] | | y' |
| [100001] | [010001] | [001001] | [000101] | [000011] | E |

The overall diffusion model $\mathcal{D}_{\epsilon,t}$ architecture is shown in Figure 5, we use a U-net architecture [57], with the approach taken for a PixelCNN++ using group normalization [58, 59]. The initial input images are progressively downsampled by a factor of 2 at each step using convolutional layers with strides of 2, progressing from $256 \times 256$ all the way down to $8 \times 8$ pixel images with swish activation functions [60]. A sinusoidal position embedding, the same as that used for Transformers [61] is used to encode the diffusion time $t$. The conditional input **c** is passed in together with the time-embedding as an additional channel at each step of the U-net. Self attention is applied at the two smallest resolution feature maps ($16 \times 16$ and $8 \times 8$).

Figure 6 gives a high level summary of the overall VAE followed by super-resolution diffusion approach, showing 2 examples of all 15 projections of a beam first generated as a 6D tensor by the VAE whose resolution is then increased by the diffusion model to capture fine details. Figure 7 shows multiple iterative denoising steps of the generative diffusion process for one example from the test data, showing all 15 generated projections alongside the true images. Figure 8 shows 9 examples of all 15 generated projections alongside the true images from the test data which shows how the beam is changing relative to modification of the solenoid magnet current. Figure 9 shows a detailed comparison of all 15 projections generated by the diffusion model relative to their true values for one test object. Figure 10 presents a quantitative summary of the ability of the diffusion process, by projecting each of the generated images onto a single axis, fitting a Gaussian to that 1D data, and then comparing those $\sigma$ fits to the true test data. This test of the accuracy of the test data ($\sigma_x, \sigma_y, \sigma_z, \sigma_{x'}, \sigma_{y'}, \sigma_E$) fits is important as these are the types of quantities that are usually measured for beams in the particle accelerator community. This last figure displays the well-known strength of diffusion-based generative models to accurately generate widely varying distributions as we see that in each case, over the entire data set where each of these $\sigma$ values spans at least 1 order of magnitude, the diffusion-generated fit have errors of at most 8%.

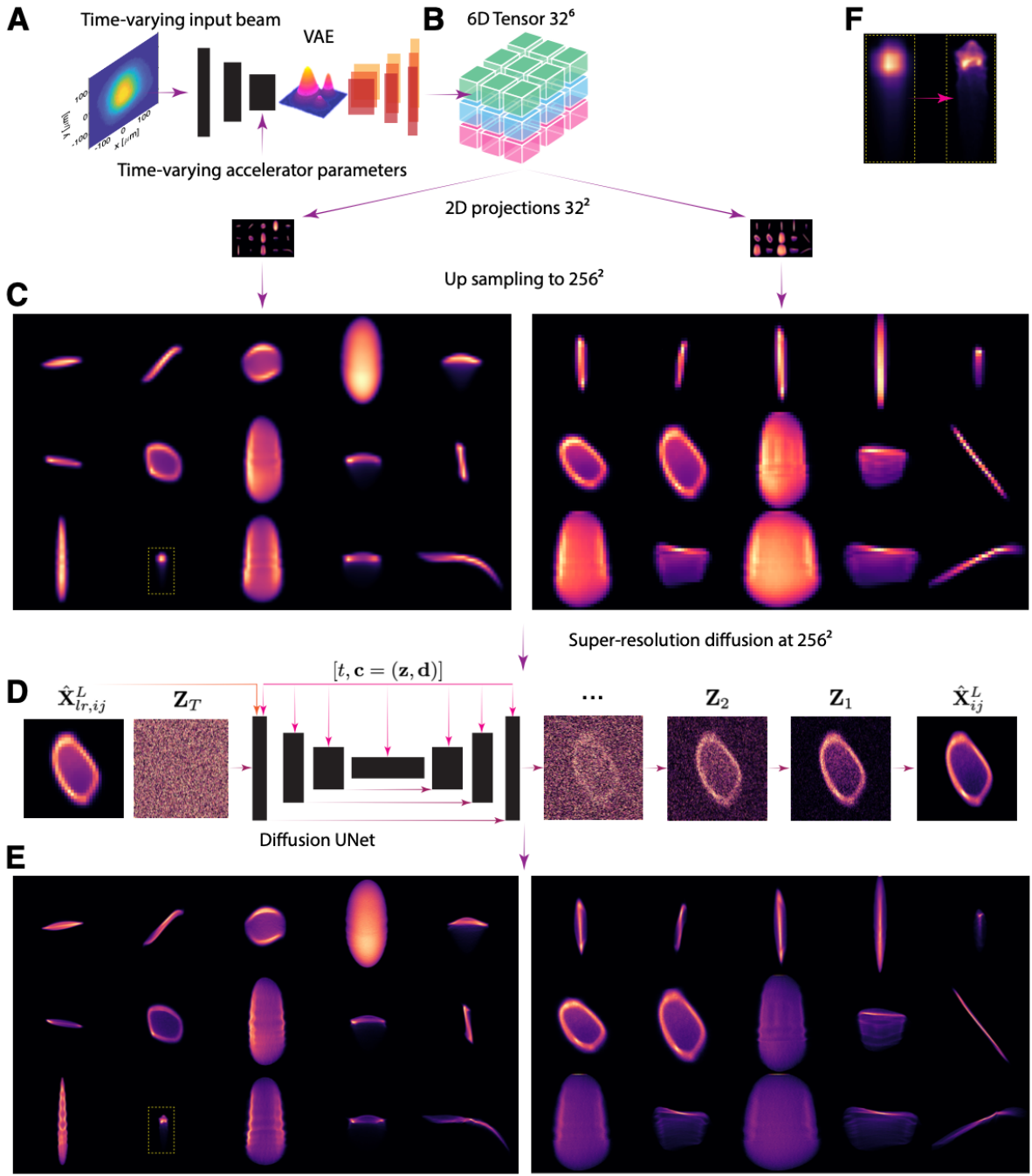

FIG. 6. Data flow through the super-resolution 6D diffusion approach. **A**: A variational autoencoder (VAE) is used to map an input beam image down to a 3 dimensional latent vector $\mathbf{z} \in \mathbb{R}^3$. This low-dimensional representation is beneficial in that it can be easily quickly adjusted to control the outputs of the decoder network using adaptive feedback. **B**: The decoder part of the VAE generates a 6D tensor of size $32^6$ which represents the beam's 6D phase space density. **C**: Projections down from the 6D tensor results in $32 \times 32$ pixel images of the 2D projections of the beam's 6D phase space density. **D**: The low resolution projections are first up-sampled to $256 \times 256$ pixels which are then used as inputs to the conditional diffusion process which is also conditioned by a conditioning vector $\mathbf{c} = [\mathbf{z}, \mathbf{d}]$ which includes the latent vector $\mathbf{z}$ and the vector $\mathbf{d} \in \mathbb{R}^6$ which defines which of the 15 2D projections to generate. **E**: By keeping $\mathbf{z}$ fixed while varying $\mathbf{d}$ all 15 high resolution projections are generated. **F**: The regions in the small yellow dashed rectangles in subplots **C** and **E** are enlarged here showing the low-resolution output on the left relative to the much more detailed super-resolution diffusion-generated output on the right.

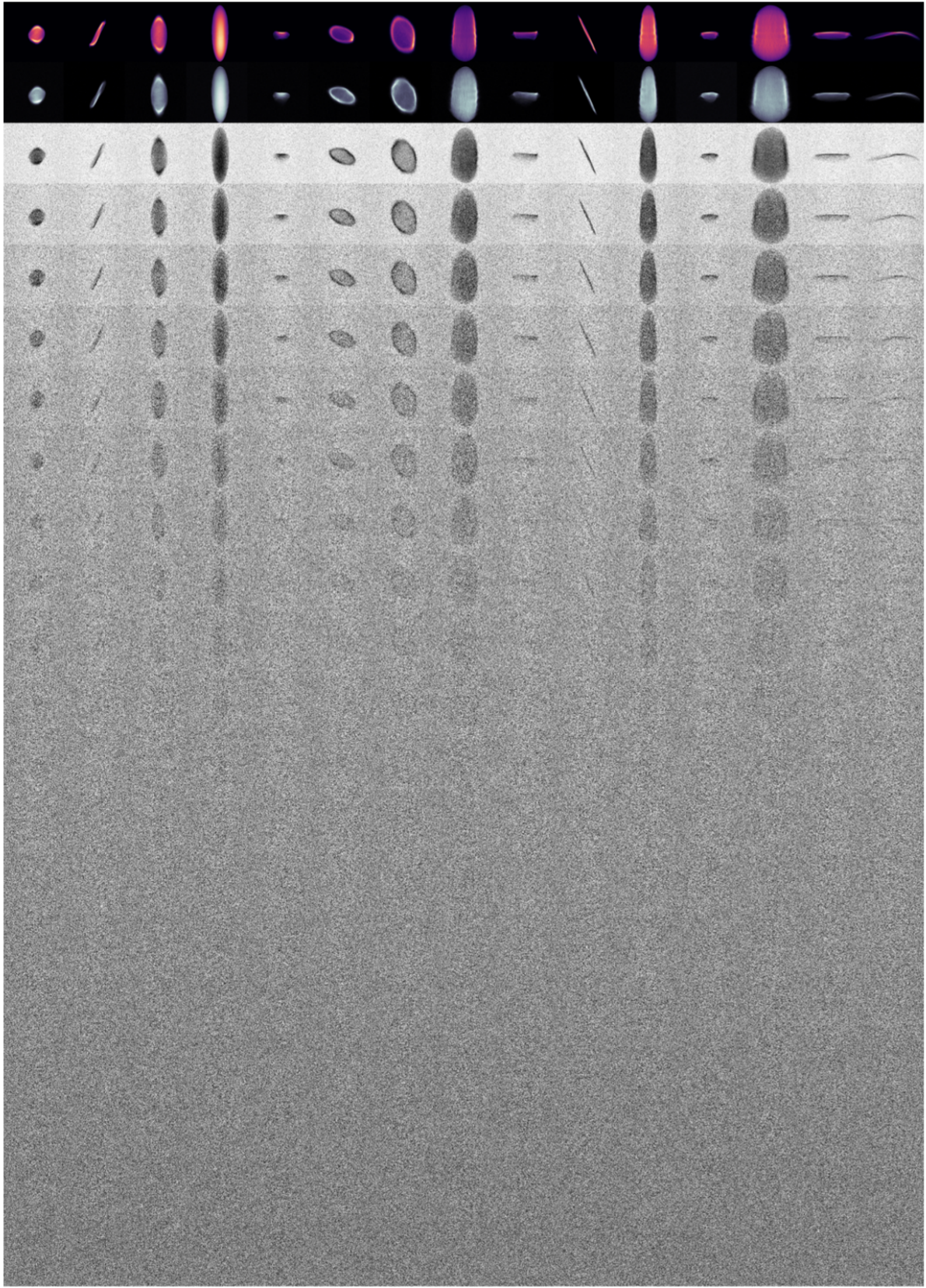

FIG. 7. Example of the generation of all 15 projections of a beam's 6D phase space distribution. The top row shows the true images, the second row is the final generated diffusion output.

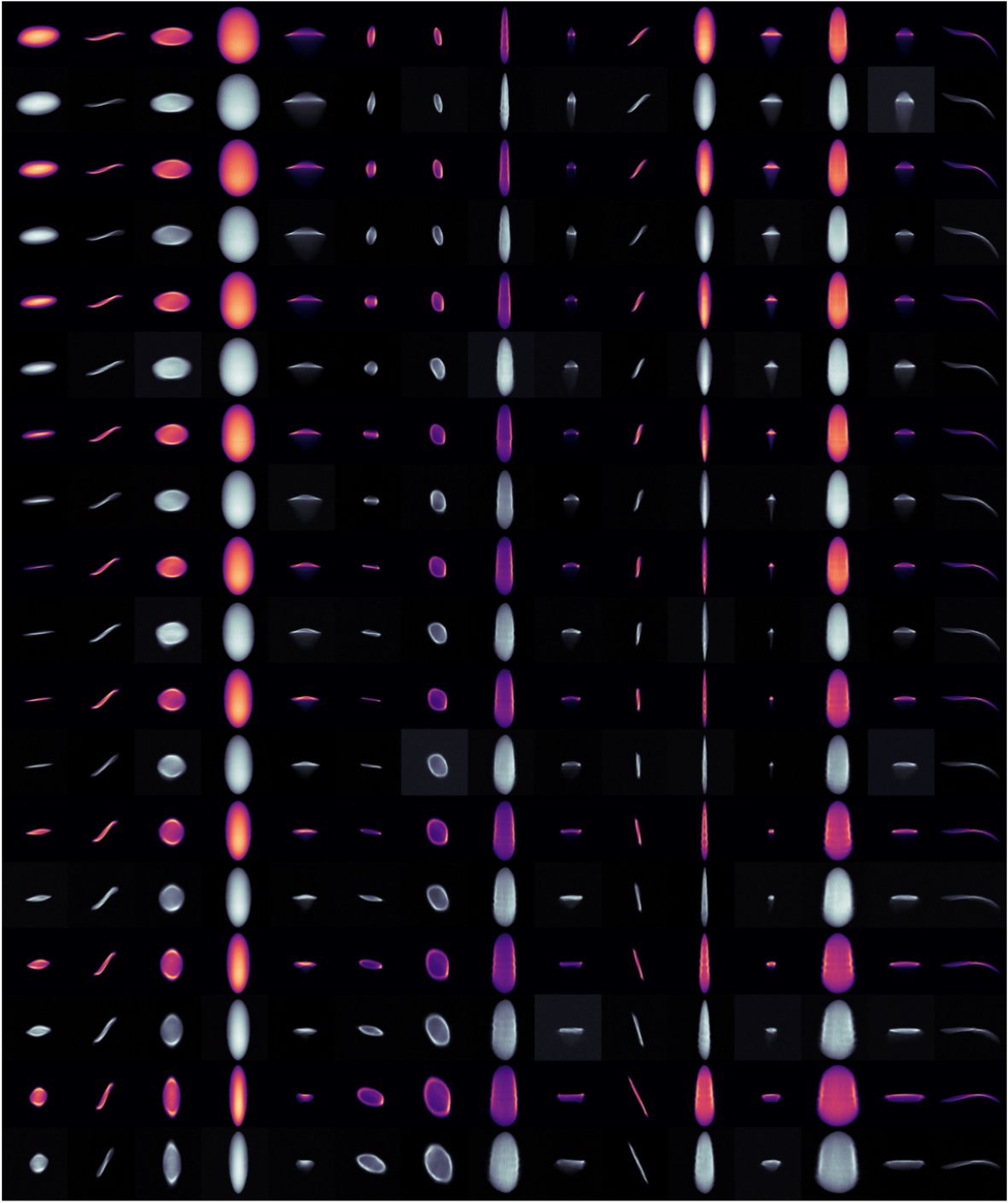

FIG. 8. Examples of the generation of all 15 projections of the 6D phase space distribution of 9 different beams compared to the true answers. Colored rows show true images, gray scale shows diffusion generated images.

## VI. ADAPTIVE LATENT SPACE TUNING

Developing ML for time varying systems with large distribution shifts is an open problem and an active field of research [62–65]. Some of the approaches to non-stationary systems rely on re-training models with new information when a system significantly changes as well as continuous learning to keep up with a system's changes [66–68]. For the case of covariate shift, where the input distribution $P(\mathbf{x})$ is different for training and test data, but the conditional distribution of output values $P(y|\mathbf{x})$ remains unchanged importance-weighting (IW) techniques have been developed [69, 70]. Kernel mean matching methods that minimize the Kullback-Leibler divergence between a test data density distribution and its estimate have also been developed [71–73]. Methods are also being developed to extracting frequency and amplitude information from time-series data [74], as well as Bayesian methods for periodic systems [75].

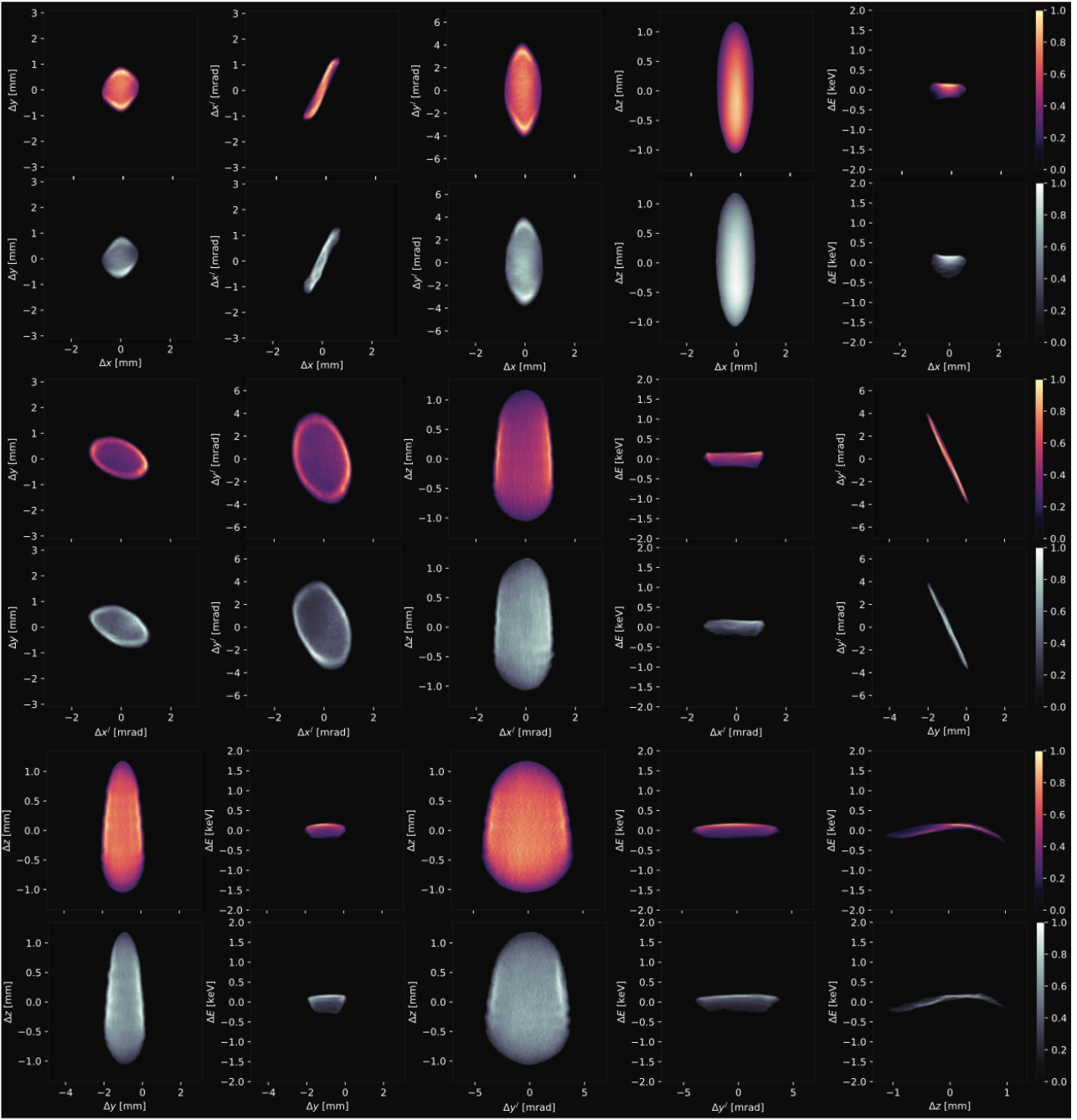

FIG. 9. One example of all 15 projections of a 6D phase space density. Reconstructed test data shown in gray compared to the true images in color.

In this work, we are focused in a combination of two forms of time variation, that of time-varying initial conditions $\mathbf{x}(t)$ and that of changing system parameters $\mathbf{p}(t)$, with the added difficulty that we do not have access to $\mathbf{x}(t)$ as that would require invasive destructive beam measurements that interupt accelerator operations. In this case, even if we could freeze the accelerator parameters at some value $\mathbf{p}(t) = \mathbf{p}(t_0)$, so that the conditional dis-tribution of output values $P(y|\mathbf{x}(t), \mathbf{p}(t_0))$ remains unchanged, we still cannot utilize the re-training methods mentioned above because we assume that we do not have access to $\mathbf{x}(t)$. Furthermore, because our parameters are time-varying $\mathbf{p}(t)$ and also assumed to be uncertain, the form of the conditional distribution of output values $P(y|\mathbf{x}(t), \mathbf{p}(t))$ is itself also a time-varying object.

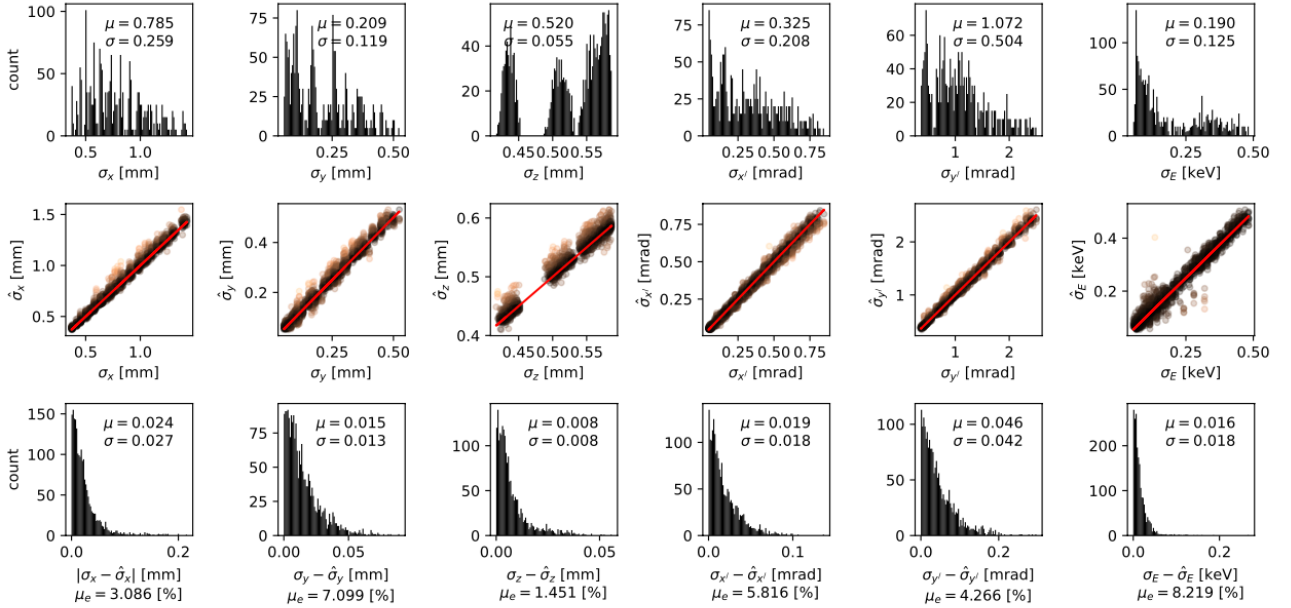

FIG. 10. Error statistics for the unseen test data. Each generated 6D distribution is projected down to one of the six axes and then fit to a Gaussian. The top row shows the standard deviation of each fit for the test data to show the wide range of the entire test data set. The middle row compares predicted standard deviations to their true values. The bottom row shows histograms of the absolute errors of the predicted standard deviations, the means and standard deviations of the error, and also the mean percent error of the predictions.

To handle such a time-varying problem we rely on robust model-independent adaptive feedback techniques that have been developed in the control theory community. Extremum seeking (ES) algorithms have been developed for the stabilization of open-loop unstable nonlinear time-varying systems [34]. We utilize a novel bounded form of ES which was developed for in-hardware applications as it has analytically guaranteed bounds on parameter updates [76], and has since been studied for a wide class of control problems [77], has been generalized to use non-differentiable dither functions [78], and has been extended to systems not affine in control [79].

The method developed in [76] is designed for time-varying nonlinear dynamic systems governing the dynamics of $\mathbf{x} \in \mathbb{R}^n$ of the form

$$\frac{d\mathbf{x}}{dt} = \mathbf{f}(\mathbf{x},t) + \mathbf{g}(\mathbf{x},t)u(\mathbf{x},t), \quad y(\mathbf{x},t) = h(\mathbf{x},t) + n(t), \tag{28}$$

where the functions $\mathbf{f}(\mathbf{x},t)$, $\mathbf{g}(\mathbf{x},t)$, and $h(\mathbf{x},t)$ are analytically unknown and we only have access to noise-corrupted measurement values $y$ of $h$ and our goal is to minimize the unknown function $h(\mathbf{x},t)$ by application of a feedback control signal $u(\mathbf{x},t)$. Note that one of the main difficulties in this case is that the unknown control direction $\mathbf{g}(\mathbf{x},t)$ can change sign, repeatedly passing through zero for functions of the form $\mathbf{g}(\mathbf{x},t) = (1+\mathbf{x})\cos(2\pi ft)$. In [76] a surprisingly simple and elegant feedback control law was designed for such systems, of the form

$$u = \sqrt{\alpha\omega}\cos\left(\omega t + ky(\mathbf{x},t)\right), \tag{29}$$

where the hyper-parameter choices for $\omega$, $\alpha$ and $k$ are made as follows. The term $k$ is a user-defined gain, for $k > 0$ the feedback performs minimization of $h$ while for $k < 0$ the function would be maximized. The dithering frequency $\omega$ should be chosen sufficiently high to change sign much faster than $\mathbf{g}(\mathbf{x},t)$, and $\alpha$ controls the dithering amplitude of $\mathbf{x}$. It can be proven that for any compact set $K \in \mathbb{R}^n$, for any $\mathbf{x}(0) \in K$, for any $T > 0$ and any desired $\delta > 0$, there exists a sufficiently large $\omega^*$ such that for all $\omega > \omega^*$ the closed loop trajectory of system (28), (29) is approximated by the averaged system dynamics

$$\frac{d\bar{\mathbf{x}}}{dt} = \mathbf{f}(\bar{\mathbf{x}},t) - \frac{k\alpha}{2}\mathbf{g}(\bar{\mathbf{x}},t)\mathbf{g}^T(\bar{\mathbf{x}},t)\nabla_{\bar{\mathbf{x}}}h(\bar{\mathbf{x}},t), \tag{30}$$

with

$$\max_{t \in [0,T]} \|\mathbf{x}(t) - \bar{\mathbf{x}}(t)\| < \delta. \tag{31}$$

Note that the main advantage of working withe the averaged dynamics (30) is that the control direction term is now positive semi-definite, of the form $\mathbf{g}\mathbf{g}^T \geq 0$, so that we are about to push $\mathbf{x}$ along gradients of the unknown function $h$ without knowing the time-varying sign of the unknown function $\mathbf{g}$.

This method is trivially extended to multi-input multi-output dynamic systems of the form

$$\frac{dx_i}{dt} = f_i(\mathbf{x}, t) + g_i(\mathbf{x}, t) u_i(\mathbf{x}, t), \quad i \in \{1, \ldots, n\}, \quad (32)$$

by utilizing controllers of the form

$$u_i = \sqrt{\alpha \omega_i} \cos(\omega_i t + ky), \quad (33)$$

where the dithering frequencies $\omega_i$ are chosen as distinct values such that $\omega_i \neq \omega_j$ for $i \neq j$, resulting in the same averaged dynamics as in (30).

In our application, to adaptively track the time-varying phase space of a beam with an unknown initial condition and unknown time-varying accelerator parameters, we rely on an ability to experimentally measure one or two of the 2D phase space projections that are being generated by the super-resolution diffusion model. In many accelerators it is possible to use a transverse deflecting RF cavity together with a dipole magnet and a scintillating screen to get single shot measurements of $(z, E)$ longitudinal phase space projections of a beam's phase space. Many times it is also possible to pass the beam directly through a scintillating screen to then measure the $(x, y)$ transverse bunch shape.

Our overall problem is then formulated as the following. We consider $\mathbf{X}^0(t)$ as the time-varying initial 6D phase space density of a beam at the accelerator entrance and $\mathbf{X}^L(t)$ as the 6D phase space density at the location $z = L$ downstream in the accelerator. We assume that parameters $\mathbf{p}(t)$, such as magnetic field strengths or bunch charge, are changing with time and so the evolution of $\mathbf{X}^0(t)$ from the accelerator entrance at $z = 0$ to the downstream location at $z = L$ is itself a nonlinear time-varying map.
so that

$$\frac{d\mathbf{X}^L(t)}{dt} = \mathbf{F}(\mathbf{X}^0(t), \mathbf{p}(t), t), \quad (34)$$

and we assume that we are able to record noisy measurements of only the $(x, y)$ and $(z, E)$ 2D projections of $\mathbf{X}^L(t)$, which we denote by

$$\mathbf{X}^L_{x,y}(t) = \Pi^{x,y} \mathbf{X}^L(t) = \iiiint \mathbf{X}^L(t) dx' dy' dz dE + n,$$

$$\mathbf{X}^L_{z,E}(t) = \Pi^{z,E} \mathbf{X}^L(t) = \iiiint \mathbf{X}^L(t) dx dy dx' dy' + n,$$

where $n$ represent random measurement noise.

For our predictions we are assuming that we do not have access to $\mathbf{X}^0(t)$ or $\mathbf{p}(t)$ and so we fix the inputs of the VAE encoder at our best guess or simply at the last available measurements, which gives us some latent embedding $\mathbf{z}_0$. We then perturb that embedding with an adaptively tuned vector $\mathbf{z}_{ES}(t)$, so that the input to the VAE's decoder is given by

$$\mathbf{z}(t) = \mathbf{z}_0 + \mathbf{z}_{ES}(t), \quad (35)$$

from which a low resolution $32^6$ pixel 6D density estimate $\hat{\mathbf{X}}^L_{lr}(t)$ is generated according to

$$\hat{\mathbf{X}}^L_{lr}(t) = F_{de}(\mathbf{z}(t)), \quad (36)$$

from which we generate low resolution $32 \times 32$ pixel estimates of the projections

$$\hat{X}^L_{lr,x,y}(t) = \Pi^{x,y} \hat{\mathbf{X}}^L_{lr}(t) = \iiiint \hat{\mathbf{X}}^L_{lr}(t) dx' dy' dz dE,$$

$$\hat{X}^L_{lr,z,E}(t) = \Pi^{z,E} \hat{\mathbf{X}}^L_{lr}(t) = \iiiint \hat{\mathbf{X}}^L_{lr}(t) dx, dy, dx' dy'. \quad (37)$$

The iterative super-resolution diffusion process guided by conditional vectors $\mathbf{c}_{x,y}(t) = [\mathbf{z}(t), \mathbf{d}_{x,y}]$ and $\mathbf{c}_{z,E}(t) = [\mathbf{z}(t), \mathbf{d}_{z,E}]$ and the interpolated high resolution image approximations $\hat{\mathbf{X}}^L_{hr,x,y}$ and $\hat{\mathbf{X}}^L_{hr,z,E}$ then generates the high resolution approximations

$$\hat{\mathbf{X}}^L_{x,y}(t) = \mathcal{D}\left[\hat{\mathbf{X}}^L_{lr,x,y}(t), \mathbf{c}_{x,y}(t)\right],$$

$$\hat{\mathbf{X}}^L_{z,E}(t) = \mathcal{D}\left[\hat{\mathbf{X}}^L_{lr,z,E}(t), \mathbf{c}_{z,E}(t)\right]. \quad (38)$$

A cost function is then defined as

$$C(\mathbf{z}(t), t) = C_{x,y}(\mathbf{z}(t), t) + C_{z,E}(\mathbf{z}(t), t), \quad (39)$$

where

$$C_{x,y}(\mathbf{z}(t), t) = \iint \left|\mathbf{X}^L_{x,y}(t) - \hat{\mathbf{X}}^L_{x,y}(t)\right| dx dy,$$

$$C_{z,E}(\mathbf{z}(t), t) = \iint \left|\mathbf{X}^L_{z,E}(t) - \hat{\mathbf{X}}^L_{z,E}(t)\right| dz dE. \quad (40)$$

Note that for notational simplicity in the above equations we have not been writing out the full dependence of the generated projections as a function of $\mathbf{z}(t)$. For example, for $\hat{X}^L_{x,y}$ this would be expressed as

$$\hat{X}^L_{x,y}(\mathbf{z}(t)) = \mathcal{D}\left[\Pi^{x,y} F_{de}(\mathbf{z}(t)), \mathbf{c}_{x,y}(t)\right]. \quad (41)$$

We then utilize the ES feedback as described above to adaptively tune the components of the latent embedding $\mathbf{z}(t)$ according to

$$\frac{dz_i(t)}{dt} = \sqrt{\alpha \omega_i} \cos\left[\omega_i t + kC(\mathbf{z}(t), t)\right], \quad (42)$$

which results in average dynamics

$$\frac{d\bar{\mathbf{z}}(t)}{dt} = -\frac{k\alpha}{2} \nabla_{\mathbf{z}} C(\mathbf{z}(t), t). \quad (43)$$

This adaptive feedback results in a time-varying predicted 6D phase space distribution

$$\frac{d\hat{\mathbf{x}}^{lr}_L(t)}{dt} = \frac{dF_{de}(\mathbf{z}(t))}{dt} = \frac{dF_{de}(\mathbf{z}(t))}{d\mathbf{z}} \frac{d\mathbf{z}}{dt}$$

$$= -\frac{k\alpha}{2} \frac{dF_{de}(\mathbf{z}(t))}{d\mathbf{z}} \nabla_{\mathbf{z}} C(\mathbf{z}(t), t), \quad (44)$$

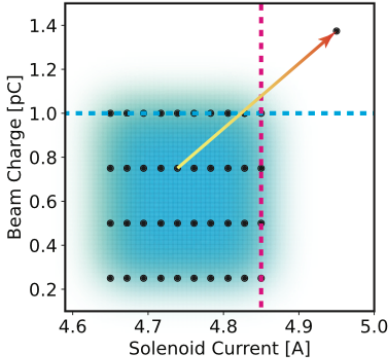

FIG. 11. The arrow shows how far the bunch charge and solenoid settings are moved beyond the training data (black dots / blue highlighted area). The color of the arrow represents the fact that the input beam initial condition is also being linearly interpolated to a measurement that was collected 6 months after the training data and therefore has never been seen by the networks.

which can be expanded as

$$-\frac{k\alpha}{2}\frac{dF_{de}(\mathbf{z}(t))}{d\mathbf{z}}\frac{dC}{d\mathcal{D}(\mathbf{z})}\left[\frac{d\Pi^{x,y}F_{de}\mathbf{z}}{d\mathbf{z}}+\frac{d\Pi^{z,E}F_{de}(\mathbf{z})}{d\mathbf{z}}\right]. \quad (45)$$

Pulling projection operators through derivatives gives

$$\frac{d\hat{\mathbf{x}}_L^{lr}(t)}{dt}=-\frac{k\alpha}{2}\left(\Pi^{z,E}+\Pi^{x,y}\right)\frac{dC}{d\mathcal{D}(\mathbf{z})}\left\|\frac{dF_{de}(\mathbf{z}(t))}{d\mathbf{z}}\right\|^2, \quad (46)$$

which highlights the strengths and limitations of the method.

First of all, an equilibrium point will be reached when $\frac{dC}{d\mathcal{D}(\mathbf{z})}=0$, that is when the cost function finds a local minimum with respect to the projections created by the generative diffusion model. A limitation of this is that the trajectory can of course get stuck at a local minimum, where the reconstruction is not accurate at all, if we start far away from the optimal solution. Another possible issue is that this minimum might not exist at all if the beam changes so much that the diffusion model cannot extrapolate in order to match it. Another possibility is that as we attempt to extrapolate we move to a region of the latent space at which $\left\|\frac{dF_{de}(\mathbf{z}(t))}{d\mathbf{z}}\right\|$ either blows up or saturates at 0, resulting in divergence from the true reconstruction or in getting stuck at an arbitrary wrong reconstruction. None of these are very surprising as this equation clearly shows what we already know, that our reconstruction ability is limited by the expressive power of both the VAE and the diffusion model.

The main benefits of the approach is that if we start near the correct solution, we are able to uniquely track the time-varying beam by this method because in this case the local minimum overlaps with the global minimum. This is again limited by the expressive power of the VAE and the diffusion networks and eventually breaks down as we move far beyond the span of the training data, but over the span of the training data we can track the beam without having access to its time-varying input or to the time-varying parameters of the accelerator in an adaptive self-supervised way.

Furthermore, as demonstrated by the subsequent results, we find that with this method we are able to extrapolate further than with previously developed adaptive approaches due to the extra robustness that we have built into the process in the form of projecting down to the individual images from a single 6D tensor which is a very strong physics constraint. This both helps to extrapolate further beyond the training set and helps to maintain physically consistent predictions even when they fail to predict the true beam. Finally, because the adaptive feedback is pushing the generated trajectories towards a minimal gradient of both the diffusion and the VAE model we find that it does not have the same kind of catastrophic failure as experienced by a model which directly takes as inputs beam and parameter measurements (if they were available). This increased robustness might also be due to not using inputs to the encoder of the VAE which are far from the training data as the adaptive setup works directly on the latent embeddings which are then passed through the decoder of the VAE only.

We demonstrate the strengths and limitations of this adaptive tracking method by starting within the span of the training data at a location where the network predictions are very accurate and then changing the true beam by using solenoid and bunch charge settings which move far beyond the span of the training data, as shown in Figure 11. As we move the accelerator parameters far beyond the range of the training data, we also linearly interpolate the experimentally measured input beam distribution from a value within the training set to one that was collected 6 months later, as represented by the color of the arrow in Figure 11.

As this change is happening we compare 3 different scenarios. For the first case, we assume that we do have access to the input beam and accelerator parameter values and use them as inputs to a trained autoencoder, as was done in [26]. In this case, the network is able to predict the correct beam distributions very accurately within the training data and then catastrophically fails as both the input beam and parameters move far away from the training values. This result is shown by projecting each of the generated 2D projections to the $(x,y,z,x',y',E)$ axes, fitting a Gaussian to each, and then plotting their mean values relative to the true value (black-dashed line) as well as the mean $\pm$ the standard deviation of the fit projections. For this first case we label the data AE and show the results in green in Figure 12. Note that because the autoencoder from [26] is generating individual images which are themselves projections of a 6D tensor as it starts to fail it quickly starts to behave in a non-physical way. For example if we focus on just the $\sigma_x$ fit the autoencoder is generating 5 different versions of this based on its generated $(x,y)$, $(x,x')$, $(x,y')$, $(x,z)$, and

$(x, E)$ distributions and the values quickly diverge.

For the second case, we show the adaptively tuned autoencoder results from [26] in which adaptive feedback is applied as described above, adjusting the latent space position in order to continuously minimize the cost function (39) by comparing predictions of the $(x, y)$ and $(z, E)$ distributions to their measurements. In this case the network does not experience catastrophic failure until much later as seen by the $\sigma_z$ and $\sigma_E$ spreads near the final steps. This is shown in red and labeled AAE in Figure 12.

Finally, for the third case we apply the adaptive latent space tuning to the VAE and then utilize the super-resolution diffusion model to create the projections that are compared to the time-varying measurements. In this case the network's predictions also start to lose accuracy as we move beyond the span of the training data, but for some of the projections the network is more accurate than in the AAE case and more importantly there is never any catastrophic failure as we have the strong built in physics constraint of first constructing all of our low-resolution images as projections of a single 6D object. This result is shown in blue and labeled diff in Figure 12.

One additional limitation of the method we are proposing which should be noted is that looking at Figure 12 we can see that when we are well within the span of the training data, the spread of the predicted $\sigma$ values shown in blue is sometimes wider than that of the adaptively tuned AE shown in red which does not project from a single 6D tensor. This can be explained by the probabilistic generative nature of the diffusion model.

In Figure 13 we show just the $(x, y)$, $(x', y')$, $(x', E)$, and $(z, E)$ projections generated by the adaptive diffusion process at steps 7 and 12 of the adaptive tracking process, alongside their low-resolution VAE-generated views. In step 7 it is difficult to see any significant difference between the true and generated images because we are still well within the range of the training data. At step 12 we are already past the boundary of training and starting to slightly extrapolate. In Figure 14 we have moved all the way to the end, to step 24 which is very far beyond the training data. The networks are now heavily extrapolating and failing to accurately capture the true data. However, even in this regime where the predictions are failing, the robustness of this adaptive physics-constrained approach which is based on projecting from a single 6D object is evident as the beam projections are physically consistent and have not exploded.

Finally, Figure 15 shows the evolution of all 15 projections and their predictions over the entire 25 step tracking process. First we see that our adaptive approach is able to track the beam very accurately during the initial steps despite having no access to the time-varying input beam or the time-varying accelerator parameters, as long as we stay within the span of the training data. Furthermore, once we go beyond the training data our approach still fights to adaptively track the unknown beam as well as possible and although its prediction accuracy begins to break down, it is still generating a physically consistent beam even when far beyond the span of its training data. The extreme difference between the various projections also once again highlights the strength of diffusion-based models to generate highly varying objects.

## VII. CONCLUSIONS

This work develops an adaptive physics-informed conditionally guided generative diffusion process. This general approach can be applied to complex dynamic systems evolving in high-dimensional phase spaces, including applications such as inverse problem solving to map backwards in time and as a virtual diagnostic for complex systems. The method has great potential for a wide range of scientific applications, especially for complex time-varying systems where there is a need for robustness and for which creating new data is computationally expensive or gathering new data experimentally requires time-consuming or invasive methods such that standard brute force re-training methods are not desirable.

For accelerator beams, the paper demonstrates that this method can be used to accurately generate unseen test data to give a non-invasive virtual high resolution view of an electron beam's longitudinal 6D phase space density. We show that the method generalizes well on unseen test data, accurately predicting complex high resolution images of a charged particle beam's phase space.

In the adaptive tracking approach for time-varying systems, we show that with adaptive feedback the method is able to track the beam very accurately despite having no access to the unknown time varying input beam at the entrance of the accelerator or to time-varying accelerator parameters. Once the beam and accelerator drift beyond the span of the training data and the model starts to extrapolate, we show that our adaptive approach is much more robust than a standard generative approach that does not incorporate adaptive feedback. Finally, even when we are far beyond the span of the training data, due to our built-in physics constraints in the form of generating and then projecting from a single 6D object, we show that our method outperforms a previously developed adaptive approach without physics constraints.

### ACKNOWLEDGMENTS

This work was funded by the US Department of Energy (DOE), Office of Science, Office of High Energy Physics under contract number 89233218CNA000001 and the Los Alamos National Laboratory LDRD Program Directed Research (DR) project 20220074DR.

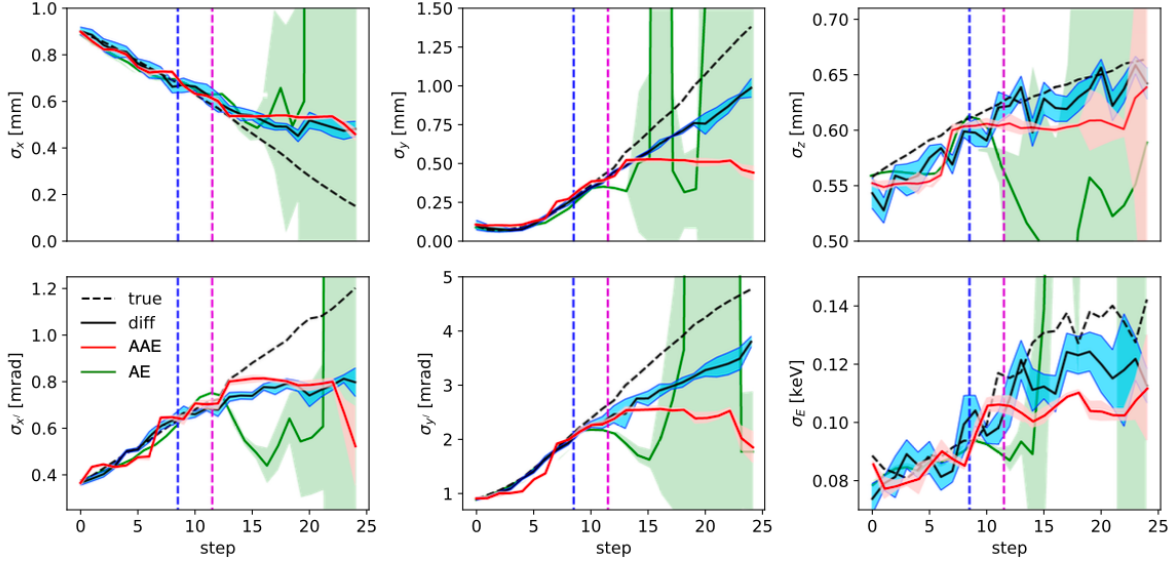

FIG. 12. Mean predictions for $(\sigma_x, \sigma_y, \sigma_z, \sigma_{x'}, \sigma_{y'}, \sigma_E)$ and the standard deviations of their spreads are compared to the true vale (black-dashed) as both the input beam and the accelerator parameters are moved far beyond the span of the training data. The method proposed in this paper, labeled diff and shown in blue, is both more accurate and more physically consistent than either of the other approaches.

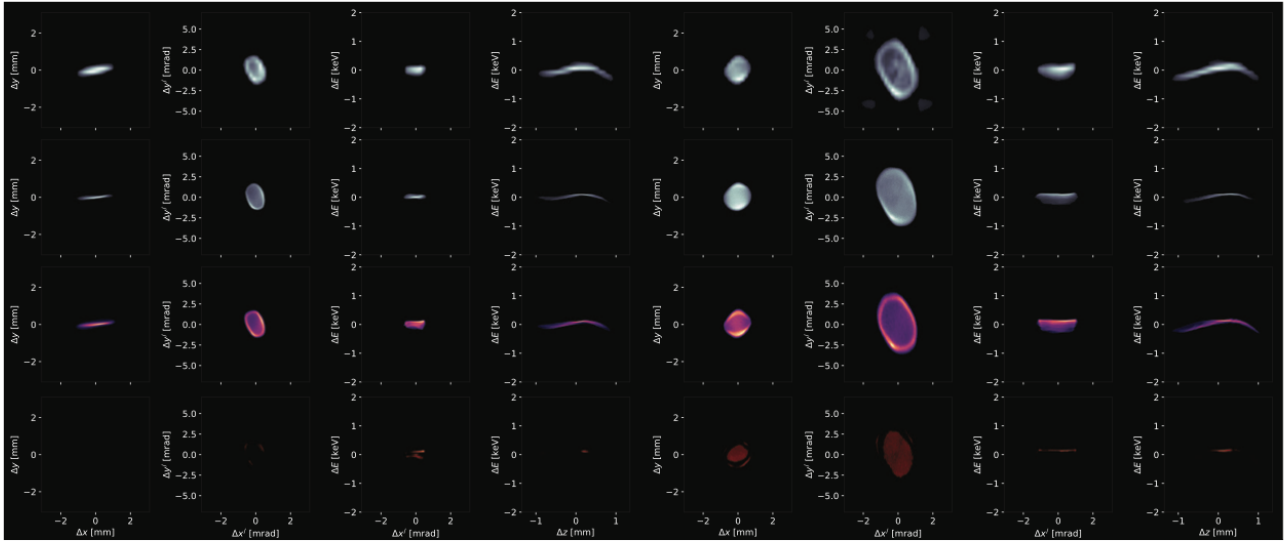

FIG. 13. Compared true and test data showing the super-resolution process for steps 7 and 12 in Figure 12. The top row shows 4 projections created directly from the 6D tensor at $32 \times 32$ resolution. The second row shows the super-resolution diffusion versions of those same projections. The third row shows the true images. The fourth row shows absolute values of differences. All images are on the same $[0, 1]$ color scales as in Figure 9.

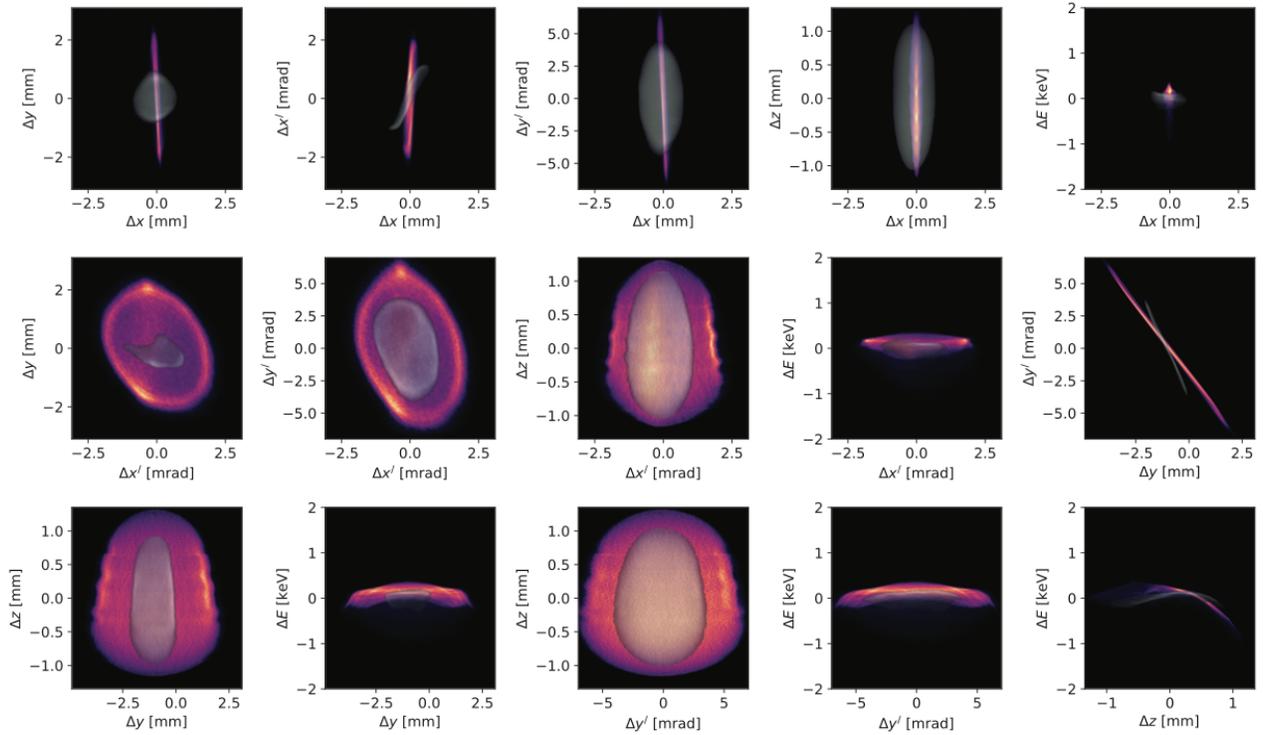

FIG. 14. At step 24 of Figure 12 the generative model is so far beyond its training data that the predictions are completely failing to track the true beam properties. The point of this figure is to show that despite the large error, the generated beam is still physically consistent as it is generated from projections of a 6D object. Here the true images are shown in color and the generated images are superimposed on top of them in semi-transparent white.

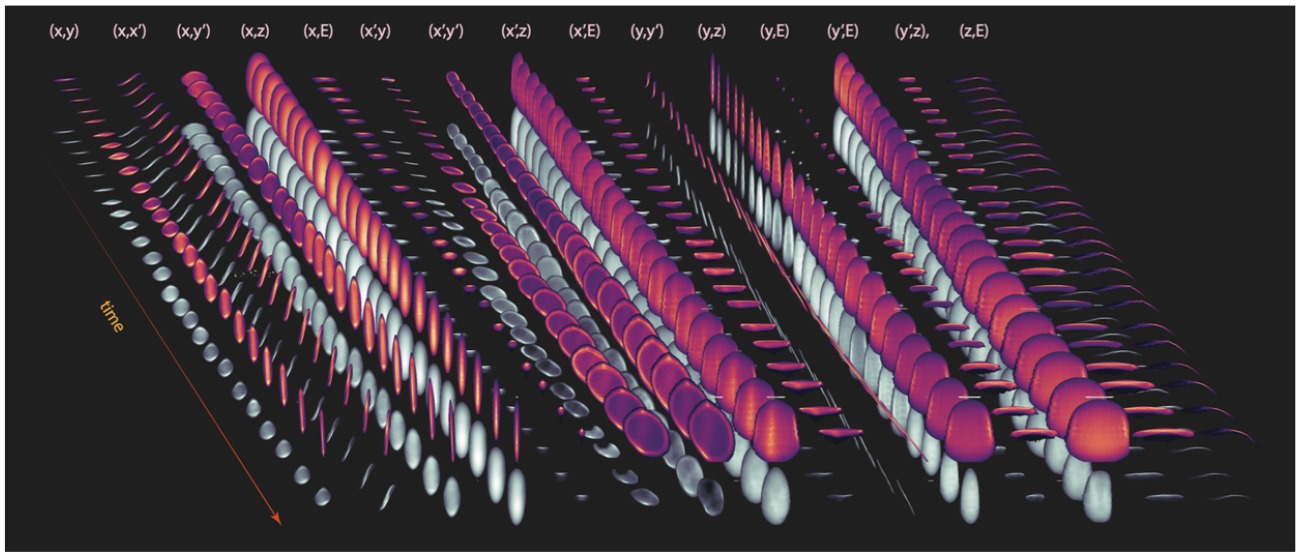

FIG. 15. First we see that our adaptive approach is able to track the beam very accurately during the initial steps despite having no access to the time-varying input beam or the time-varying accelerator parameters, as long as we stay within the span of the training data. Furthermore, once we go beyond the training data our approach still fights to adaptively track the unknown beam as well as possible and although its prediction accuracy begins to break down, it is still generating a physically consistent beam even when far beyond the span of its training data.